\documentclass[10pt,twocolumn,letterpaper]{article}

\usepackage{cvpr}              

\usepackage[dvipsnames]{xcolor}

\usepackage{multirow}
\usepackage{setspace}

\definecolor{cvprblue}{rgb}{0.21,0.49,0.74}
\usepackage[pagebackref,breaklinks,colorlinks,citecolor=cvprblue]{hyperref}

\def\paperID{37} 
\def\confName{3DV\xspace}
\def\confYear{2024\xspace}

\title{MonoLSS: Learnable Sample Selection For Monocular 3D Detection}

\author{Zhenjia Li\thanks{Equal contribution.}\\
  Baidu Inc.\\
  Beijing, China \\
{\tt\small lizhenjia@baidu.com}
\and
Jinrang Jia\footnotemark[1]\\
  Baidu Inc.\\
  Beijing, China \\
{\tt\small jiajinrang@baidu.com}
\and
  Yifeng Shi\thanks{Corresponding author.}\\
  Baidu Inc.\\
  Beijing, China \\
  {\tt\small shiyifeng@baidu.com} \\
}

\begin{document}
\maketitle
\begin{abstract}
In the field of autonomous driving, monocular 3D detection is a critical task which estimates 3D properties (depth, dimension, and orientation) of objects in a single RGB image. Previous works have used features in a heuristic way to learn 3D properties, without considering that inappropriate features could have adverse effects. In this paper, sample selection is introduced that only suitable samples should be trained to regress the 3D properties. To select samples adaptively, we propose a Learnable Sample Selection (LSS) module, which is based on Gumbel-Softmax and a relative-distance sample divider. The LSS module works under a warm-up strategy leading to an improvement in training stability. Additionally, since the LSS module dedicated to 3D property sample selection relies on object-level features, we further develop a data augmentation method named MixUp3D to enrich 3D property samples which conforms to imaging principles without introducing ambiguity. As two orthogonal methods, the LSS module and MixUp3D can be utilized independently or in conjunction. Leveraging the LSS module and the MixUp3D, without any extra data, our method named MonoLSS ranks \textbf{1st} in all three categories (Car, Cyclist, and Pedestrian) on KITTI 3D object detection benchmark, and achieves competitive results on both the Waymo dataset and KITTI-nuScenes cross-dataset evaluation. The code is available at https://github.com/Traffic-X/MonoLSS.
\end{abstract}    
\section{Introduction}
\label{sec:intro}
3D object detection has received increasing attention in the fields of autonomous driving \cite{10.1145/3460426.3463656, Lang_2019_CVPR, Li_2022_CVPR}, intelligent traffic \cite{Yu_2023_CVPR, jia2023competition, Yu_2022_CVPR}, and robot navigation. Compared to expensive LIDAR sensor \cite{Chen_2023_ICCV, Wu_2022_CVPR,Yang_2019_ICCV, 10.1145/3581783.3611948}, monocular camera which enables a greater range of environmental perception and object localization is economical. In recent years, many visual monocular 3D detection algorithms \cite{peng2022did,liu2022monocon,ye2020monocular, liu2021ground, kumar2021groomed, jia2023monouni} have been proposed to improve perception effect. Different from 2D object detection, accurate 3D property predictions (depth, dimension, and orientation), especially depth \cite{Lian_2022_CVPR,Ma_2021_CVPR}, are more critical and challenging in this field.

\begin{figure}[t]
\begin{center}
\includegraphics[width=1.0\linewidth]{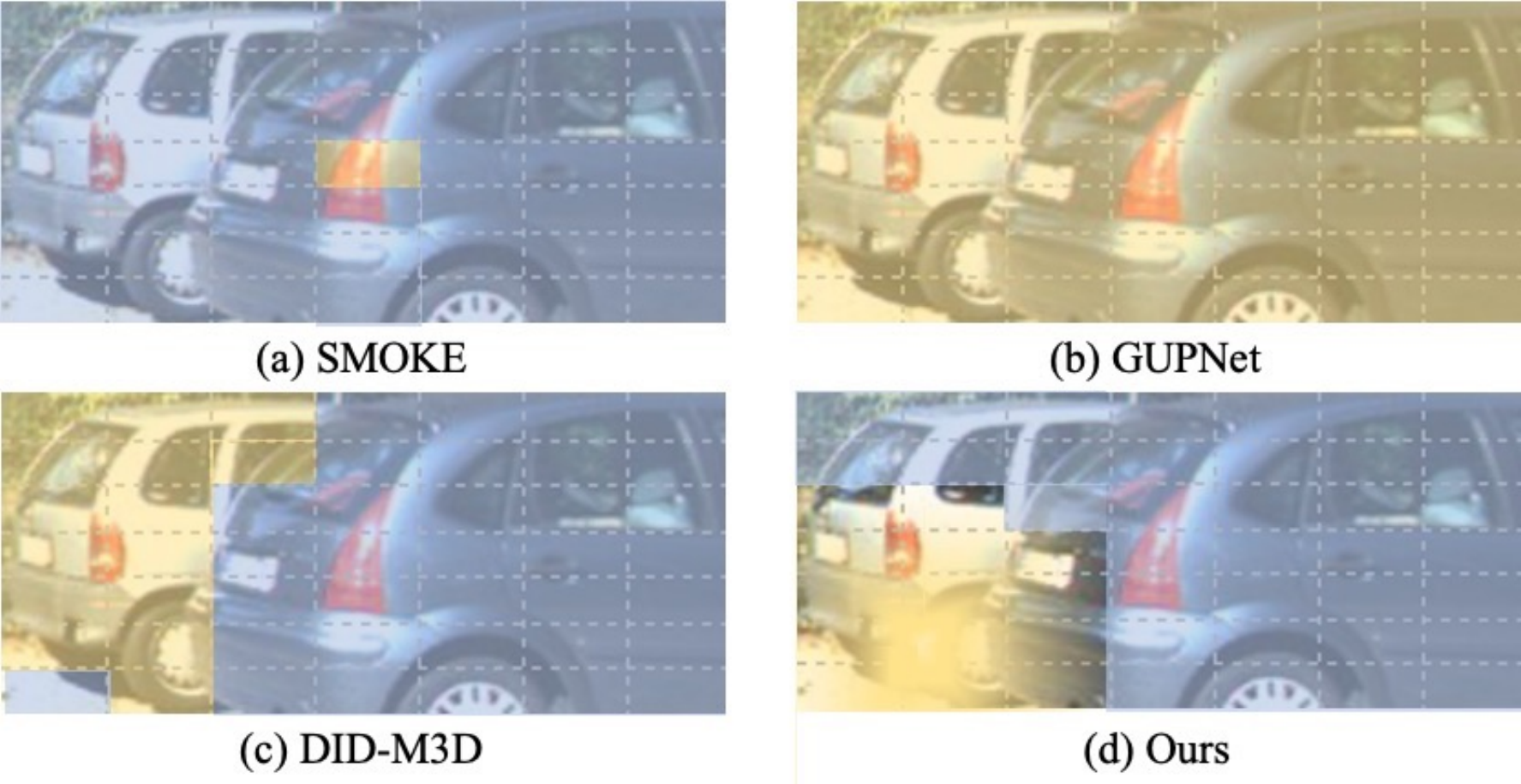}
\end{center}
\vspace{-15pt}
\caption{\textbf{Comparison among various features using by different methods for 3D property learning.} When the occluded white vehicle is the target of detection, various methods utilize distinct features for learning. Yellow color means features used and blue means not.}
\vspace{-5pt}
\label{fig1_compare}
\end{figure}

To achieve accurate 3D property estimations, many methods add 3D property prediction branches into 2D detectors \cite{2017Faster,zhou2019objects}. These branches leverage the features extracted by backbone networks to output 3D properties. However, it should be noted that not all of the features are appropriate for learning 3D properties. The design motivation comes from the label assignment of 2D detection. People rarely require an achor with IOU less than 0.3 as a positive sample for target detection (in anchor free method, it means far from object center). This is because the visual features do not match the learning objectives. The use of inappropriate ones can result in ambiguity and even have adverse effects. We transfer this knowledge to 3D property learning. For example, as shown in Figure \ref{fig1_compare}, the white car which has a feature map with size $d * d * C$ is occluded by a gray one. SMOKE \cite{Liu_2020_CVPR_Workshops} only uses one fixed-position feature with size $1 * 1 * C$ located on the 3D center of object to regress 3D properties. When occlusion occurs, this feature may lie on another object. Although the receptive field is not limited to the location of the feature, the network may not receive the optimal information as input. In contrast, GUPNet \cite{Lu_2021_ICCV} takes advantage of all the $d * d$ features and outputs 3D properties with a global average pooling module \cite{Lin2013NetworkIN}. Suffering from useless information including foreground and background interference, this approach remains problematic.

In this work, we introduce sample selection to identify the features that are beneficial for learning 3D properties and serve as positive samples, while disregarding the rest and treating them as negative samples. The challenge lies in how to divide them. An intuitive approach is to focus on the features of the target objects themselves (Figure \ref{fig1_compare} (c)), but these methods require the introduction of additional data such as depth maps \cite{peng2022did} or segmentation labels, and still cannot select suitable samples among different internal components of objects, such as wheels, lights, or bodies. To address the 3D property sample selection problem, we propose a novel Learnable Sample Selection (LSS) module. The LSS module implements probability sampling with Gumbel-Softmax \cite{jang2017categorical}. Furthermore, top-k Gumbel-Softmax \cite{kool2019stochastic} is employed to enable multi-sample sampling, expanding the number of samples drawn from 1 to $k$. Moreover, to replace the use of a same $k$ value for all objects, we developed a hyperparameter-free sample divider based on relative distance, which achieves adaptive determination of sampling values for each object. Additionally, inspired by HTL method \cite{Lu_2021_ICCV}, the LSS module works with a warm-up strategy to stabilize training process.

Furthermore, the LSS module dedicated to 3D property sample selection relies on object-level features. However, the object number in training data is always limited. Meanwhile, most 3D monocular data augmentation methods, such as random crop-expand, random flip, copy and paste, etc., do not change the features of objects themselves. Some of them even introduce ambiguous features due to the violation of imaging principles. In order to improve the richness of the 3D property samples, we propose MixUp3D, which adds physical constraints on the basis of traditional 2D MixUp \cite{zhang2018mixup} to simulate spatial overlap in the physical world. The spatial overlap does not change 3D properties of the objects, such as a car overlapping a bicycle, but we can still judge their depths, dimensions, and orientations. As a simulation of the spatial overlap, the MixUp3D enables the objects to conform to imaging principles without introducing ambiguity. It can enrich training samples and alleviate overfitting. Moreover, the MixUp3D can be used as a fundamental data augmentation method in any monocular 3D detection approach.

Incorporating all the techniques, our monocular 3D detection method named MonoLSS outperforms prior state-of-the-art (SOTA) works with a significant margin without using any extra data. It can be trained end-to-end simply while still maintaining real-time efficiency. To summarize, the main contributions of this work are as follows:

\begin{itemize}
\item We emphasize that not all features are equally effective for learning 3D properties, and first reformulate it as a problem of sample selection. Correspondingly, a novel Learnable Sample Selection (LSS) module that can adaptively select samples is developed.
\item To enrich 3D property samples, we devise MixUp3D data augmentation, which simulates spatial overlap and improves 3D detection performance.
\item Without introducing any extra information, MonoLSS ranks 1st in all the three classes in the KITTI benchmark \cite{geiger2012we} and surpasses the current best method by more than 11.73$\%$ and 12.19$\%$ relatively on the Moderate and Hard levels of the Car class. It also achieves SOTA results on Waymo dataset \cite{sun2020scalability} and KITTI-nuScenes \cite{caesar2020nuscenes} cross-dataset evaluation.
\end{itemize}
\section{Related work}
\label{sec:rela}

\noindent
\textbf{Monocular 3D Object Detection.} The monocular 3D object detection aims to predict accurate 3D bounding boxes. According to whether use extra data, monocular 3D object detection algorithms can be mainly categorized into two groups. The first kind of method only uses a single image as input without any extra information. For example, M3D-RPN \cite{Brazil_2019_ICCV} adopts a standalone 3D region proposal network and proposes a depth-wise convolution to predict objects. Based on a CenterNet-style \cite{zhou2019objects} network, SMOKE \cite{Liu_2020_CVPR_Workshops} predicts 3D bounding boxes by combining a single keypoint estimation module. Furthermore, MonoFlex \cite{Zhang_2021_CVPR} optimizes the truncated obstacles prediction method with an edge heatmap and edge fusion module. MonoPair \cite{Chen_2020_CVPR} explores the relationships between different objects. MonoEF \cite{Zhou_2021_CVPR} first predicts camera extrinsic parameters by detecting vanishing point and horizon change, and then adopts a converter to rectify perturbative features in the latent space. MonoCon \cite{liu2022monocon} learns auxiliary monocular contexts projected from the 3D bounding boxes in training and discards them for better inference efficiency in inference. MonoDDE \cite{Li_2022_CVPR} exploits depth clues in monocular images and develops a model which produces 20 depths for each target.

\begin{figure*}
\centering
\includegraphics[width=175mm]{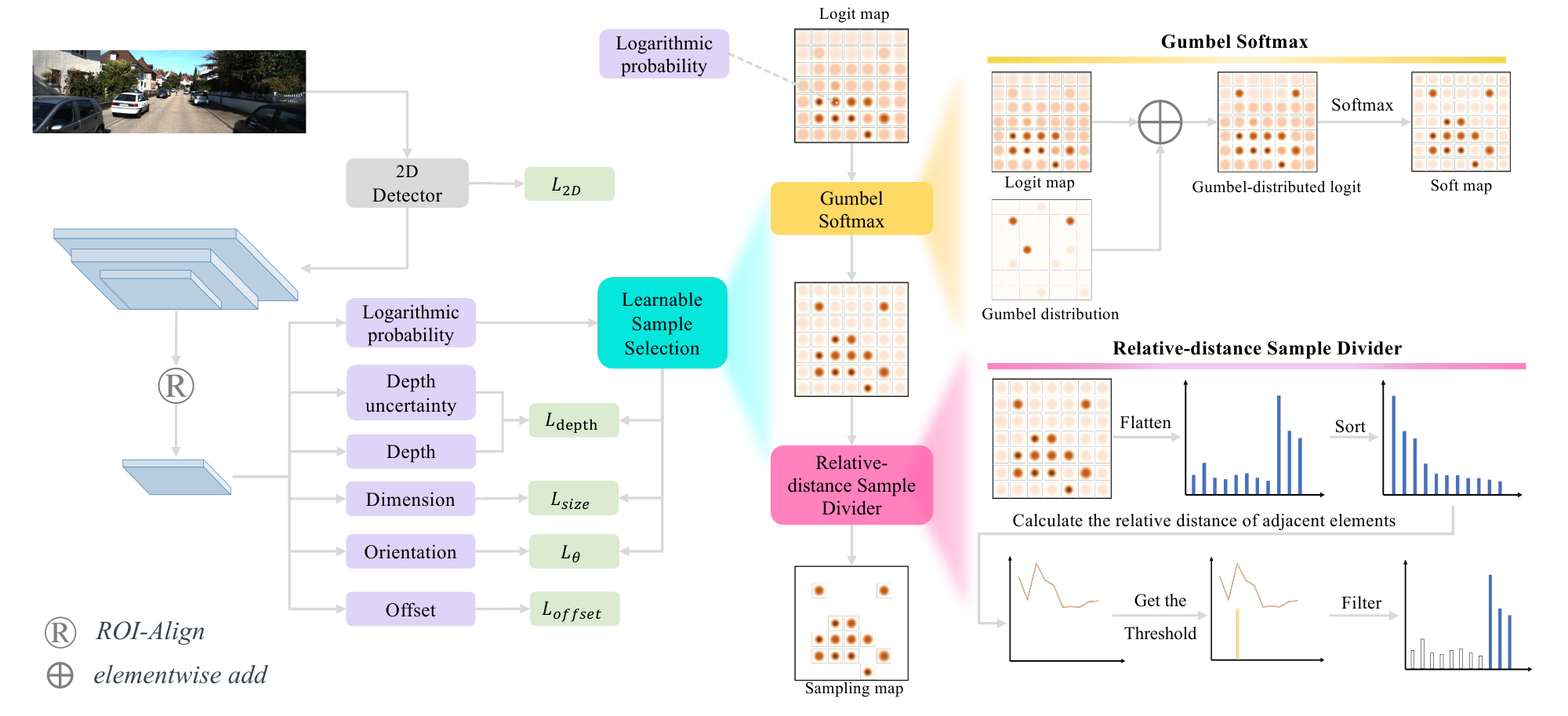}
\vspace{-15pt}
\caption{\textbf{An overview of the MonoLSS framework.} First, a 2D detector combined with ROI-Align is used to generate object features. Then, six heads respectively predict 3D properties (depth, dimension, orientation, and 3D center projection offset), depth uncertainty, and logarithmic probability. Finally, the Learnable Sample Selection (LSS) module adaptively selects samples and acts on the loss calculation.}
\vspace{-5pt}
\label{overview}
\end{figure*}

The second kind of method uses extra data, such as depth maps, LIDAR point clouds and CAD models, to obtain additional information, and enhance detection. ROI-10D \cite{Manhardt_2019_CVPR} combines the deep feature maps and estimates dense depth maps to regress 3D bounding boxes. D4LCN \cite{Ding_2020_CVPR} proposes depth-guided convolution in which the receptive field is determined adaptively by the predicted depth. DID-M3D \cite{peng2022did} decouples the instance depths into attribute depths and visual depths by using a dense depth map. CaDDN \cite{Reading_2021_CVPR} uses LIDAR points to generate depth maps and estimates depth by an additional monocular network, then converts the feature to BEV perspective for prediction. CMKD \cite{YuH-CMKD-ECCV2022} develops a cross-modality method to transfer the knowledge from LiDAR modality to image. Besides depth maps and LIDAR, methods such as AutoShape \cite{Liu_2021_ICCV} utilize CAD models to generate dense keypoints to alleviate the sparse constraints. 

Methods that leverage additional data always exhibit superior performance due to the increased information. However, the complex sensor configuration and computational overhead limit their practical applications in industry.

\noindent
\textbf{Sample Selection in 2D/3D Detection.} According to how to allocate samples, 2D object detection methods can be mainly categorized into two groups. The anchor-based methods \cite{Deng_2020_CVPR,Redmon_2017_CVPR,2017Faster} allocate positive samples based on the Intersection Over Union (IOU) between target boxes and pre-defined anchors, while the anchor-free methods \cite {10.1007/978-3-031-18916-6_41} are based on certain rules. ATSS \cite{Zhang_2020_CVPR} allocates samples by setting adaptive IOU thresholds based on statistical characteristics of the object. MTL \cite{Ke_2020_CVPR} finds the best sample points by gradually reducing the number of positive samples.

The sample assign strategy for 2D properties of 3D detection methods generally follows those mentioned above. Many methods \cite{Li_2022_CVPR,Liu_2020_CVPR_Workshops,Zhang_2021_CVPR} use these strategies consistent with 2D properties to learn 3D properties. Methods \cite{Lu_2021_ICCV,Shi_2021_ICCV} use object features extract from backbone by ROI-Align \cite{2017Faster} to regress one 3D property, which leads the results suffer from the foreground and background interference. DID-M3D \cite{peng2022did} uses dense depth maps to select positive samples, which requires extra annotations. 

\noindent
\textbf{Data Augmentation in Monocular 3D Detection.}
Due to the violation of geometric constraints, random horizontal flipping \cite{Chen_2020_CVPR,Li_2022_CVPR,Zou_2021_ICCV} and photometric distortion \cite{Chen_2021_CVPR,liu2022monocon} are the only two data augmentation methods mostly used in monocular 3D detection. Some methods \cite{peng2022did} use random crop and expand to simulate the proportional change of depth. However, according to the imaging principles, it is impractical for all depths on one image to have the same proportional change. Some methods \cite{Lian_2022_CVPR, Ye_2022_CVPR} use an additional depth map to simulate the forward and backward movement of the camera along the z-axis. While, due to parallax and depth map errors, this methods introduce a lot of noise and distorted appearance features. Instance-level copy-paste \cite{Lian_2022_CVPR} is also used as a 3D data augmentation method, but limited by the complex manual processing logic, it is still not realistic enough.
\section{ Methodology}

Monocular 3D object detection extracts features from a single RGB image, estimates the category and 3D bounding box for each object in the image. The 3D bounding box can be further divided into 3D center location ($x$, $y$, $z$), dimension ($h$, $w$, $l$) and orientation (yaw angle) $\theta$. The roll and pitch angles of objects are set to 0.

In this work, we propose a novel Learnable Sample Selection (LSS) module to optimize the monocular 3D object detection process. The overall architecture of the MonoLSS is illustrated in Figure \ref{overview}, which mainly includes 2D detector, ROI-Align, 3D detection heads, and LSS module.

Our 2D detector is built on CenterNet \cite{zhou2019objects}. It takes an image $ I\in\mathbb{R}^{H\times W\times3}$ as input and adopts DLA34 \cite{Yu_2018_CVPR} to compute the deep feature $F\in\mathbb{R}^{\frac{H}{4}\times \frac{W}{4}\times C}$, where $C$ is the channel number. Then deep feature $F$ is fed into three 2D detection heads to regress 2D offset, 2D size and 2D heatmap. We achieve 2D boxes by combining these 2D predictions and further use ROI-Align to extract object features $F_{obj}\in\mathbb{R}^{n\times d \times d\times C}$ from deep feature $F$, where $d \times d$ is the ROI-Align size and $n$ refers to the number of ROIs.

Unlike other methods that only predict a single 3D bounding box based on object features, our method uses each sample point in object features to predict a 3D bounding box and a logarithmic probability. In addition, we follow the multi-bin design for predicting the orientation and predict the uncertainty for depth, which is the same as GUPNet \cite{Lu_2021_ICCV}. Therefore, we have 3D box dimension $S_{3d}\in\mathbb{R}^{n\times d\times d\times3}$, 3D center projection offset $O_{3d}\in\mathbb{R}^{n\times2}$, orientation $\Theta\in\mathbb{R}^{n \times d\times d\times 12\times2}$, depth $D\in\mathbb{R}^{n\times d\times d}$, depth uncertainty $U\in\mathbb{R}^{n\times d\times d}$, and logit map $\Phi\in\mathbb{R}^{n\times d\times d}$. Based on the logit map predicted by the network, the LSS module can adaptively select positive samples for 3D properties when training. During inference, the LSS module selects the best 3D properties according to the highest logarithmic probability in logit map.

\subsection{Learnable Sample Selection}
Assert $U \sim Uniform(0,1)$, then we can use inverse transform sampling to generate the Gumbel distribution $G$ by computing $G = -log(-log(U))$. By independently perturbing the log-probabilities with Gumbel distribution and using the $argmax$ function to find the largest element, the Gumbel-Max trick \cite{gumbel1954statistical} achieves probability sampling without random choices. Based on this work, Gumbel-Softmax \cite{jang2017categorical} uses the softmax function as the continuous, differentiable approximation to $argmax$, and achieves overall differentiability with the help of reparameterization. Gumbel-Top-k \cite{kool2019stochastic} extends the number of sample points from Top-1 to Top-k by drawing an ordered sampling of size $k$ without replacement, where the $k$ is a hyperparameter. However, a same $k$ is not suitable for all objects, for instance, occluded objects should have fewer positive samples than normal ones. To this end, we design a hyperparameter-free relative-distance based module to divide samples adaptively. In summary, we propose a Learnable Sample Selection (LSS) module to address sample selection problem in 3D property learning, which is formed by Gumbel-Softmax and relative-distance sample divider. The diagram of the LSS module is shown on the right side of Figure \ref{overview}.

Let $\Phi = \{ \phi_1, \phi_2, ... , \phi_N \}$ be the logit map output by the model, where $N=d \times d$ denotes the number of sample points. Each element $\phi_i$ for $i \in [1, N]$ represents a logarithmic probability. Gumbel-Softmax is performed to the logit map $\Phi$ to achieve probability sampling. Concretely, we first generate the Gumbel distribution $G$ with size $ d \times d $ based on the previous description and add it to the logit map $\Phi$ to obtain the Gumbel-distributed logit $\hat{\Phi} = \{ \hat{\phi}_1, \hat{\phi}_2, ..., \hat{\phi}_N \}$. In this way, the originally low logarithmic probability values have the opportunity to surpass other high logarithmic probability values. Then a temperature-scaled softmax is adopted to process $\hat{\Phi}$ to get soft map $S = \{ S_1, S_2, ... , S_N\}$. The overall process can be formulated as follows:
\begin{align}
    \hat{\Phi} = (G&+\Phi)  \\
    S_i = \frac{exp(\hat{\phi}_i/\tau))}{\sum_{j=1}^{N}{exp(\hat{\phi}_j/\tau)}} &\  for \ i = 1,...,N
\end{align}
where the temperature coefficient $\tau$ is set to 1 in this work.

After that, the relative-distance sample divider is adopted to replace the fixed $k$ in Gumbel-Top-k to implement adaptive sample allocation. We use the maximum interval between the elements of the soft map to distinguish positive and negative samples. Generally, absolute distance $(Abs\_dis = |a - b|)$ is used to indicate the interval. However, due to the amplification effect of the softmax function, using the absolute distance to divide samples may lead to an insufficient number of positive samples. We employ the relative distance $(Rel\_dis = \frac{a}{b})$ to increase the number of positive samples\footnote{See the Supplementary Material for more proof details.}. For example, if the softmax function has an input vector [20, 18, 17, 7], the output would be [0.84, 0.12, 0.04, 0], using the absolute distance will assign only one positive sample while the relative distance assigns three. 

First, we flatten the soft map $S$ to a one-dimensional vector $Soft\_S$ and sort it to get a sorted vector $Sort\_S$. Second, we calculate the relative distance between adjacent elements of the vector $Sort\_S$ by the following formula:

\begin{equation}
\begin{aligned}
    &Dis\_S_i = \frac{Sort\_S_i}{Sort\_S_{i+1}} = \frac{exp(\hat{\phi}_{f(i)}/\tau)}{exp(\hat{\phi}_{f(i+1)}/\tau)} \\ &=exp(\frac{\hat{\phi}_{f(i)} - \hat{\phi}_{f(i+1)}}{\tau}) \  for \ i = 1,...,N-1
\end{aligned}
\end{equation}
where $f()$ denotes the mapping relation from $Sort\_S$ to $Soft\_S$. Assume the $Dis\_S_i$ is the maximum value in $Dis\_S$, since the exponential function is a monotone increasing function, the $\hat{\phi}_{f(i)} - \hat{\phi}_{f(i+1)}$ is the maximum interval in $\hat{\Phi}$. In essence, we are finding the most discriminative value in Gumbel-distributed logit $\hat{\Phi}$ to distinguish between positive and negative samples. We choose the $Sort\_S_i$ corresponding to the $Dis\_S_i$ as the threshold to filter negative samples. Specifically, the values in the soft map $S$ that smaller than $Sort\_S_i$ are set to 0 to obtain the final sampling map $Sample\_S$. Consequently, reparameterization \cite{jang2017categorical} trick is performed to $Sample\_S$ to achieve differentiability. Based on the above designs, the LSS module realizes the derivable dynamic sample allocation without hyperparameters.

\subsection{Loss Function and Training Strategy}
The overall loss $L$ consists of 2D loss $L_{2d}$ and 3D loss$L_{3d}$, where the 2D loss $L_{2d}$ follows the design in CenterNet \cite{zhou2019objects} and the 3D loss $L_{3d}$ employs a multi-task loss function based on the LSS module to supervise the learning of the 3D properties:
\begin{align}
    L = L_{2d} + L_{O_{3d}} + (L_{S_{3d}}  + L_{depth} + L_{\theta} ) \cdot Sample\_S
\end{align}
where the $L_{S_{3d}}$ is L1 loss for dimensions predict, and the $L_{O_{3d}}$ denotes Smooth-L1 loss for 3D center projection offset regression. The $L_{depth}$ denotes the depth loss where we employ the Laplacian aleatoric uncertainty loss \cite{Chen_2020_CVPR,kendall2017uncertainties} to supervise depth estimation. The $L_{\theta}$ denotes orientation loss, which use the multi-bin loss that follow the \cite{Mousavian_2017_CVPR}. After calculating the loss of each property, we multiply loss map of the depth, orientation and dimension properties by final sampling map $Sample\_S$ of LSS to prevent the loss backpropagation of the negative samples.

The HTL \cite{Lu_2021_ICCV} training strategy is employed to reduce instability. In addition, since the instability of the 3D properties loss will interfere with the selection of positive sample points, we use a warm-up strategy when training the LSS module. Specifically, all samples will be used to learn 3D properties in the early stage of training until the depth loss stabilizes. We consider the depth loss stabilization as a necessary condition to start the LSS module, since depth is the most important property that affects the accuracy of 3D bounding boxes \cite{Lian_2022_CVPR,Ma_2021_CVPR}.

\begin{figure}[t]
\begin{center}
\includegraphics[width=1.0\linewidth]{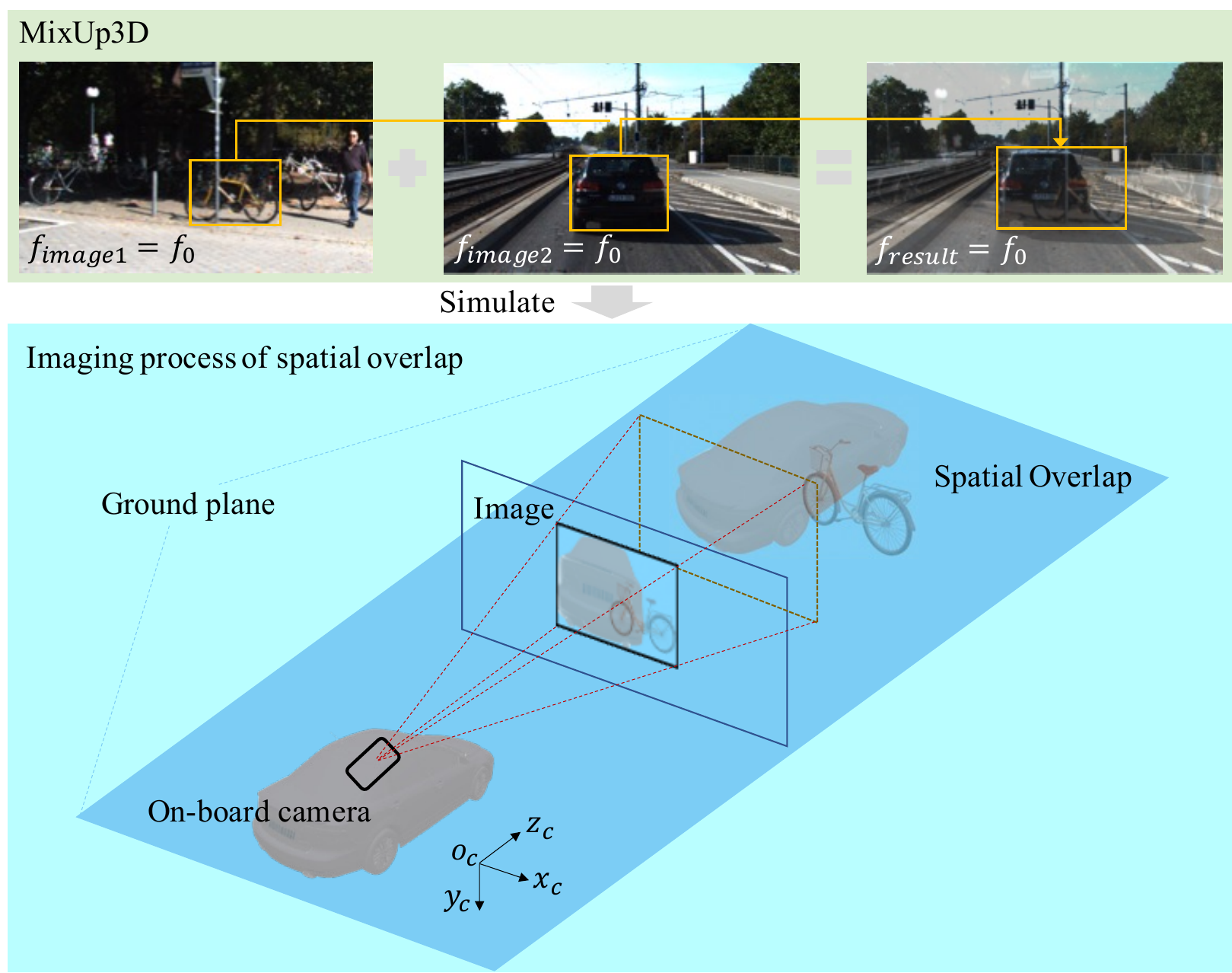}
\end{center}
\vspace{-15pt}
   \caption{\textbf{Visualization of the MixUp3D which simulates spatial overlap.} A car overlaps a bicycle in the physical world and their appearance features in resulting image do not introduce ambiguity for 3D property learning.}
   \vspace{-5pt}
\label{MixUp}
\label{fig:long}
\label{fig:onecol}
\end{figure}

\subsection{MixUp3D for Spatial Overlap Simulation}
Due to strict imaging constraints, data augmentation methods are limited in monocular 3D detection. Besides photometric distortion and horizontal flipping, most data augmentation methods introduce ambiguous features due to breaking the imaging principles. Additionally, since the LSS module focuses on object-level features, methods that do not modify the features of the objects themselves are not expected to be effective enough for the LSS module.

Thanks to the advantages of the MixUp \cite{zhang2018mixup,zhang2020does}, pixel-level features of objects can be enhanced. We propose MixUp3D, which adds physical constraints to the 2D MixUp, enabling the newly generated image is essentially plausible imaging of spatial overlap. Specifically, the MixUp3D violates only the collision constraints of objects in the physical world, while ensuring that the resulting image adheres to imaging principles, thus avoiding any ambiguity.

Traditional MixUp method blends different images proportionally in a 2D pixel coordinate system, without considering whether the resulting image is compatible with the imaging principles in the 3D physical world. For example, two images with different focal lengths or resolutions are always directly mixed together, which introduces depth ambiguity. Another example is that the mixing of images taken from different views can result in confusion with regard to perspective. In this paper, we impose strict constraints on the MixUp images to ensure they have the same focal length, principal points, resolution and views of camera (pitch and roll angle). This enables the simulation of an image captured by one camera at a single time with spatial overlap by leveraging images taken by two cameras at different times and locations. Generally, images with the same focal length always mean that their principal points and resolutions are also the same. Concurrently, the images are all captured by on-board pinhole cameras, with their $x_c$-axis and $z_c$-axis being parallel to the ground plane, leading to similar views. Therefore, the MixUp3D only needs to consider ensuring that the focal lengths of the images are the same. The schematic diagram of the MixUp3D is shown in Figure \ref{MixUp}. 

\begin{table*}
\footnotesize
\setstretch{0.9}
\begin{center}
\tabcolsep=0.023\linewidth
\begin{tabular}{l|c|c|ccc|ccc|c}
\hline
\multirow{2}*{Method} & \multirow{2}*{Reference} & \multirow{2}*{Extra Data} & \multicolumn{3}{c|}{$AP_{3D}(IOU=0.7|{R_{40}})$} & \multicolumn{3}{c|}{$AP_{BEV}(IOU=0.7|{R_{40}})$} & Runtime \\ \cline{4-9}
 & & & Easy & \textbf{Mod.} & Hard & Easy & Mod. & Hard & (ms) \\
\hline\hline
MonoPSR \cite{Ku_2019_CVPR} & CVPR 2019 & LIDAR & 10.76 & 7.25 & 5.85 & 18.33 & 12.58 & 9.91 & 200 \\
PatchNet \cite{Ma_2020_ECCV} & ECCV 2020 & Depth & 15.68 & 11.12 & 10.17 & 22.97 & 16.86 & 14.97 & 400 \\
MonoRUn \cite{Chen_2021_CVPR} & CVPR 2021 & LIDAR & 19.65 & 12.30 & 10.58 & 27.94 & 17.34 & 15.24 & 70 \\
CaDDN \cite{Reading_2021_CVPR} & CVPR 2021 & LIDAR & 19.17 & 13.41 & 11.46 & 27.94 & 18.91 & 17.19 & 630 \\
DFR-Net \cite{Zou_2021_ICCV} & ICCV 2021 & Depth & 19.40 & 13.63 & 10.35 & 28.17 & 19.17 & 14.84 & 180 \\
AutoShape \cite{Liu_2021_ICCV} & ICCV 2021 & CAD & 22.47 & 14.17 & 11.36 & 30.66 & 20.08 & 15.59 & 50 \\
DID-M3D \cite{peng2022did} & ECCV 2022 & Depth & 24.40 & 16.29 & 13.75 & 32.95 & 22.76 & 19.83 & 40 \\
DD3D \cite{Park_2021_ICCV} & ICCV 2021 & Depth & 23.22 & 16.34 & 14.20 & 30.98 & 22.56 & 20.03 & - \\
CMKD \cite{YuH-CMKD-ECCV2022} & ECCV 2022 & LIDAR & 25.09 & 16.99 & 15.30 & 33.69 & 23.10 & 20.67 & - \\
\hline
M3D-RPN \cite{Brazil_2019_ICCV} & ICCV 2019 & None & 14.76 & 9.71 & 7.42 & 21.02 & 13.67 & 10.23 & 160 \\
SMOKE \cite{Liu_2020_CVPR_Workshops} & CVPR 2020 & None & 14.03 & 9.76 & 7.84 & 20.83 & 14.49 & 12.75 & 30 \\
MonoPair \cite{Chen_2020_CVPR} & CVPR2020 & None & 13.04 & 9.99 & 8.65 & 19.28 & 14.83 & 12.89 & 60 \\
MonoDLE \cite{Ma_2021_CVPR} & CVPR2021 & None & 17.23 & 12.26 & 10.29 & 24.79 & 18.89 & 16.00 & 40 \\
MonoFlex \cite{Zhang_2021_CVPR} & CVPR 2021 & None & 19.94 & 13.89 & 12.07 & 28.23 & 19.75 & 16.89 & 30 \\
GUPNet \cite{Lu_2021_ICCV} & ICCV 2021 & None & 20.11 & 14.20 & 11.77 & - & - & - & 30 \\
DEVIANT \cite{kumar2022deviant} & ECCV 2022 & None & 21.88 & 14.46 & 11.89 & 29.65 & 20.44 & 17.43 & - \\
MonoCon \cite{liu2022monocon} & AAAI 2022 & None & 22.50 & 16.46 & 13.95 & 31.12 & 22.10 & 19.00 & 25 \\
MonoDDE \cite{Li_2022_CVPR} & CVPR 2022 & None & \underline{24.93} & 17.14 & \underline{15.10} & 33.58 & 23.46 & 20.37 & 40 \\
\hline
\textbf{MonoLSS (Ours)} & - & None & \textbf{26.11} & \textbf{19.15} & \textbf{16.94} & \textbf{34.89} & \textbf{25.95} & \textbf{22.59} & 35 \\
\hline
\end{tabular}
\end{center}
\vspace{-15pt}
\caption{\textbf{Monocular 3D detection performance of Car category on KITTI \textit{test} set.} All results are evaluated on KITTI testing server. Same as KITTI leaderboard, methods are ranked under the moderate difficulty level. We highlight the best results in bold and the second ones in underlined. For the extra data: 1) \textbf{LIDAR} denotes methods use extra LIDAR cloud points in training process. 2) \textbf{Depth} means utilizing depth maps or models pre-trained under another depth estimation dataset. 3) \textbf{CAD} denotes using dense shape annotations provided by CAD models. 4) \textbf{None} means no extra data is used.}
\label{table_1}
\end{table*}

\begin{table}
\begin{center}
\footnotesize
\setstretch{0.9}
\tabcolsep=0.46em
\begin{tabular}{l|c|ccc|ccc}
\hline
\multirow{3}*{Method} & \multirow{3}*{Extra} & \multicolumn{6}{c}{$AP_{3D}(IOU=0.5|{R_{40}})$} \\
\cline{3-8}
 & & \multicolumn{3}{c}{Pedestrian} & \multicolumn{3}{|c}{Cyclist} \\
 \cline{3-8}
 & & Easy & \textbf{Mod.} & Hard & Easy & Mod. & Hard \\
\hline\hline
DFR-Net \cite{Zou_2021_ICCV} & Depth & 6.09 & 3.62 & 3.39 & 5.69 & 3.58 & 3.10 \\
CaDDN \cite{Reading_2021_CVPR} & LIDAR & 12.87 & 8.14 & 6.76 & 7.00 & 3.41 & 3.30 \\
DD3D \cite{Park_2021_ICCV} & Depth & 13.91 & 9.30 & 8.05 & 2.39 & 1.52 & 1.31 \\
CMKD \cite{YuH-CMKD-ECCV2022} & LIDAR & 17.79 & 11.69 & 10.09 & 9.60 & 5.24 & 4.50 \\ 
\hline
MonoDDE \cite{Li_2022_CVPR} & None & 11.13 & 7.32 & 6.67 & \underline{5.94} & \underline{3.78} & \underline{3.33} \\
MonoCon \cite{liu2022monocon} & None & 13.10 & 8.41 & 6.94 & 2.80 & 1.92 & 1.55 \\
GUPNet \cite{Lu_2021_ICCV} & None & 14.95 & 9.76 & 8.41 & 5.58 & 3.21 & 2.66 \\
MonoDTR \cite{Huang_2022_CVPR} & None & \underline{15.33} & \underline{10.18} & \underline{8.61} & 5.05 & 3.27 & 3.19 \\
\hline
\textbf{MonoLSS} & None & \textbf{17.09} & \textbf{11.27} & \textbf{10.00} & \textbf{7.23} & \textbf{4.34} & \textbf{3.92} \\
\hline
\end{tabular}
\end{center}
\vspace{-15pt}
\caption{\textbf{Monocular 3D detection performance of Pedestrian and Cyclist category on KITTI \textit{test} set.}}
\label{table_2}
\end{table}

Considering a training dataset $I=\{I_{f_1}, I_{f_2}, ..., I_{f_K}\}$, where $I_{f_k}$ for $k \in [1, K]$ means images in $I_{f_k}$ have the same focus $f_k$. We denote images and corresponding labels as $I_{f_k} = \{(x^{f_k}_1, y^{f_k}_1), (x^{f_k}_2, y^{f_k}_2), ..., (x^{f_k}_N, y^{f_k}_N)\}$, which has totally $N$ samples. The MixUp3D process can be defined as the following form:
\begin{equation}
\left\{
\begin{array}{l}
x^{f_k}_n = \lambda \cdot x^{f_k}_n + (1 - \lambda) \cdot x^{f_k}_m \\
y^{f_k}_n = y^{f_k}_n + y^{f_k}_m
\end{array}
\right.
\end{equation}
where $n, m \in [1, N]$ and $n \neq m$. $\lambda$ denotes mix proportion. 

The proposed MixUp3D can enrich training samples without introducing ambiguity, and effectively alleviate overfitting problems. It can be conveniently applied in any monocular 3D detection task as an essential data augmentation method.
\section{Experiments}
\subsection{Setup}
\noindent
\textbf{Dataset and Evaluation metrics.} We evaluate our proposed method on the widely used KITTI \cite{geiger2012we}, Waymo \cite{sun2020scalability} and nuScenes \cite{caesar2020nuscenes} benchmarks.
\begin{itemize}
\item \textbf{\textit{KITTI}} consists of 7481 training images and 7518 testing images. It has three classes (Car, Pedestrian, and Cyclist), each with three difficulty levels (Easy, Moderate, and Hard).  We follow the prior work \cite{chen20153d} to divide the 7481 training images into a training set (3712) and validation set (3769) for ablation study. Following the official protocol \cite{Simonelli_2019_ICCV}, we use $AP_{3D|R_{40}}$ and $AP_{BEV|R_{40}}$ on Moderate category as main metrics.
\item \textbf{\textit{Waymo}} evaluates objects at two levels: Level 1 and Level 2, based on the number of LiDAR points present in their 3D box. The evaluation is conducted at three distances: [0, 30), [30, 50), and [50, $\infty$) meters. Waymo utilizes the $APH_{3D}$ percentage metric, which incorporates heading information in $AP_{3D}$, as a benchmark for evaluation.
\item \textbf{\textit{nuScenes}} comprises 28,130 training and 6,019 validation images captured from the front camera. We use validation split for cross-dataset evaluation.
\end{itemize}

\noindent
\textbf{Implementation details.} Our proposed MonoLSS is trained on 4 Tesla V100 GPUs with a batch size of 16. Without the MixUp3D, we train the model for 150 epochs, after which overfitting occurrs. While using MixUp3D, the model can be trained for 600 epochs without overfitting. We use Adam \cite{kingma2014adam} as our optimizer with an initial learning rate $1e-3$. The learning scheduler has a linear warm-up strategy in the first 5 epochs. Following \cite{peng2022did}, the ROI-Align size $d \times d $ is set to $7\times7$. The LSS starts after 0.3 $\times$ total epochs (experimental parameters) for warmup.

\begin{table*}
\footnotesize
\setstretch{0.95}
\begin{center}
\tabcolsep=0.0165\linewidth
\begin{tabular}{c|c|l|c|cccc|cccc}
\hline
\multirow{2}*{$IOU_{3D}$} & \multirow{2}*{Difficulty} & \multirow{2}*{Method} & \multirow{2}*{Extra} & \multicolumn{4}{c|}{$AP_{3D}$} & \multicolumn{4}{c}{$APH_{3D}$} \\ \cline{5-12}
 & & & & All & 0-30 & 30-50 & 50-$\infty$ & All & 0-30 & 30-50 & 50-$\infty$ \\
\hline\hline
\multirow{7}*{0.7} & \multirow{7}*{Level\_1} & CaDDN \cite{Reading_2021_CVPR} & LIDAR & 5.03 & 15.54 & 1.47 & 0.10 & 4.99 & 14.43 & 1.45 & 0.10 \\
 &  & PatchNet \cite{Ma_2020_ECCV} in \cite{wang2021progressive} & Depth & 0.39 & 1.67 & 0.13 & 0.03 & 0.39 & 1.63 & 0.12 & 0.03 \\
 &  & PCT \cite{wang2021progressive} & Depth & 0.89 & 3.18 & 0.27 & 0.07 & 0.88 & 3.15 & 0.27 & 0.07 \\
 &  & M3D-RPN \cite{Brazil_2019_ICCV} in \cite{Reading_2021_CVPR} & None & 0.35 & 1.12 & 0.18 & 0.02 & 0.34 & 1.10 & 0.18 & 0.02 \\
 &  & GUPNet \cite{Lu_2021_ICCV} in \cite{kumar2022deviant} & None & 2.28 & 6.15 & 0.81 & \underline{0.03} & 2.27 & 6.11 & 0.80 & \underline{0.03} \\
 &  & DEVIANT \cite{kumar2022deviant} & None & \underline{2.69} & \underline{6.95} & \underline{0.99} & 0.02 & \underline{2.67} & \underline{6.90} & \underline{0.98} & 0.02 \\
 &  & \textbf{MonoLSS (Ours)} & None & \textbf{3.71} & \textbf{9.82} & \textbf{1.14} & \textbf{0.16} & \textbf{3.69} & \textbf{9.75} & \textbf{1.13} & \textbf{0.16} \\
\hline
\multirow{7}*{0.7} & \multirow{7}*{Level\_2} & CaDDN \cite{Reading_2021_CVPR} & LIDAR & 4.49 & 14.50 & 1.42 & 0.09 & 4.45 & 14.38 & 1.41 & 0.09 \\
 &  & PatchNet \cite{Ma_2020_ECCV} in \cite{wang2021progressive} & Depth & 0.38 & 1.67 & 0.13 & 0.03 & 0.36 & 1.63 & 0.11 & 0.03 \\
 &  & PCT \cite{wang2021progressive} & Depth & 0.66 & 3.18 & 0.27 & 0.07 & 0.66 & 3.15 & 0.26 & 0.07 \\
 &  & M3D-RPN \cite{Brazil_2019_ICCV} in \cite{Reading_2021_CVPR} & None & 0.35 & 1.12 & 0.18 & 0.02 & 0.33 & 1.10 & 0.17 & 0.02 \\
 &  & GUPNet \cite{Lu_2021_ICCV} in \cite{kumar2022deviant} & None & 2.14 & 6.13 & 0.78 & 0.02 & 2.12 & 6.08 & 0.77 & 0.02 \\
 &  & DEVIANT \cite{kumar2022deviant} & None & \underline{2.52} & \underline{6.93} & \underline{0.95} & \underline{0.02} & \underline{2.50} & \underline{6.87} & \underline{0.94} & \underline{0.02} \\
 &  & \textbf{MonoLSS (Ours)} & None & \textbf{3.27} & \textbf{9.79} & \textbf{1.11} & \textbf{0.15} & \textbf{3.25} & \textbf{9.73} & \textbf{1.10} & \textbf{0.15} \\
\hline
\multirow{7}*{0.5} & \multirow{7}*{Level\_1} & CaDDN \cite{Reading_2021_CVPR} & LIDAR & 17.54 & 45.00 & 9.24 & 0.64 & 17.31 & 44.46 & 9.11 & 0.62 \\
 &  & PatchNet \cite{Ma_2020_ECCV} in \cite{wang2021progressive} & Depth & 2.92 & 10.03 & 1.09 & 0.23 & 2.74 & 9.75 & 0.96 & 0.18 \\
 &  & PCT \cite{wang2021progressive} & Depth & 4.20 & 14.70 & 1.78 & 0.39 & 4.15 & 14.54 & 1.75 & 0.39 \\
 &  & M3D-RPN \cite{Brazil_2019_ICCV} in \cite{Reading_2021_CVPR} & None & 3.79 & 11.14 & 2.16 & \underline{0.26} & 3.63 & 10.70 & 2.09 & 0.21 \\
 &  & GUPNet \cite{Lu_2021_ICCV} in \cite{kumar2022deviant} & None & 10.02 & 24.78 & 4.84 & 0.22 & 9.94 & 24.59 & 4.78 & \underline{0.22} \\
 &  & DEVIANT \cite{kumar2022deviant} & None & \underline{10.98} & \underline{26.85} & \underline{5.13} & 0.18 & \underline{10.89} & \underline{26.64} & \underline{5.08} & 0.18 \\
 &  & \textbf{MonoLSS (Ours)} & None & \textbf{13.49} & \textbf{33.64} & \textbf{6.45} & \textbf{1.29} & \textbf{13.38} & \textbf{33.39} & \textbf{6.40} & \textbf{1.26} \\
\hline
\multirow{7}*{0.5} & \multirow{7}*{Level\_2} & CaDDN \cite{Reading_2021_CVPR} & LIDAR & 16.51 & 44.87 & 8.99 & 0.58 & 16.28 & 44.33 & 8.86 & 0.55 \\
 &  & PatchNet \cite{Ma_2020_ECCV} in \cite{wang2021progressive} & Depth & 2.42 & 10.01 & 1.07 & 0.22 & 2.28 & 9.73 & 0.97 & 0.16 \\
 &  & PCT \cite{wang2021progressive} & Depth & 4.03 & 14.67 & 1.74 & 0.36 & 4.15 & 14.51 & 1.71 & 0.35 \\
 &  & M3D-RPN \cite{Brazil_2019_ICCV} in \cite{Reading_2021_CVPR} & None & 3.61 & 11.12 & 2.12 & \underline{0.24} & 3.46 & 10.67 & 2.04 & \underline{0.20} \\
 &  & GUPNet \cite{Lu_2021_ICCV} in \cite{kumar2022deviant} & None & 9.39 & 24.69 & 4.67 & 0.19 & 9.31 & 24.50 & 4.62 & 0.19 \\
 &  & DEVIANT \cite{kumar2022deviant} & None & \underline{10.29} & \underline{26.75} & \underline{4.95} & 0.16 & \underline{10.20} & \underline{26.54} & \underline{4.90} & 0.16 \\
 &  & \textbf{MonoLSS (Ours)} & None & \textbf{13.12} & \textbf{33.56} & \textbf{6.28} & \textbf{1.15} & \textbf{13.02} & \textbf{33.32} & \textbf{6.22} & \textbf{1.13} \\
\hline
\end{tabular}
\end{center}
\vspace{-15pt}
\caption{\textbf{Monocular 3D detection performance of Vehicle category on Waymo \textit{val} set.}}
\vspace{-5pt}
\label{table_waymo}
\end{table*}


\begin{table}
\begin{center}
\footnotesize
\setstretch{0.9}
\tabcolsep=0.35em
\begin{tabular}{l|cccc|cccc}
\hline
\multirow{2}*{Method} & \multicolumn{4}{c}{KITTI Val} & \multicolumn{4}{|c}{nuScenes frontal Val} \\
\cline{2-9}
 & 0-20 & 20-40 & 40-$\infty$& All & 0-20 & 20-40 & 40-$\infty$ & All \\
\hline\hline
M3D-RPN \cite{Brazil_2019_ICCV} & 0.56 & 1.33 & 2.73 & 1.26 & 0.94 & 3.06 & 10.36 & 2.67 \\
MonoRCNN \cite{shi2021geometry} & 0.46 & 1.27 & 2.59 & 1.14 & 0.94 & 2.84 & 8.65 & 2.39 \\
GUPNet \cite{Lu_2021_ICCV} & 0.45 & 1.10 & 1.85 & 0.89 & 0.82 & \underline{1.70} & 6.20 & 1.45 \\
DEVIANT \cite{kumar2022deviant} & 0.40 & 1.09 & 1.80 & 0.87 & \underline{0.76} & \textbf{1.60} & \textbf{4.50} & \textbf{1.26} \\
\hline
\textbf{MonoLSS} & \textbf{0.35} & \textbf{0.89} & \textbf{1.77} & \textbf{0.82} & \textbf{0.59} & 2.01 & \underline{5.40} & \underline{1.42} \\
\hline
\end{tabular}
\end{center}
\vspace{-15pt}
\caption{\textbf{Cross-dataset evaluation of the KITTI \textit{val} model on KITTI \textit{val} and nuScenes frontal \textit{val} cars with depth MAE.}}
\vspace{-5pt}
\label{table_cross}
\end{table}

\subsection{Main Results}
\noindent
\textbf{Results of Car category on KITTI test set.} As shown in Table \ref{table_1}, our proposed MonoLSS achieves superior performance than previous methods, even those with extra data. Specifically, compared with MonoDDE \cite{Li_2022_CVPR} which is the recent top1-ranked image-only method, MonoLSS gains significant improvement of \textbf{4.73\%/11.73\%/12.19\%} in $AP_{3D}$ and \textbf{3.90\%/10.61\%/10.90\%} in $AP_{BEV}$ relatively on the easy, moderate, and hard levels while $IOU=0.7$. 

\noindent
\textbf{Results of Pedestrian/Cyclist on KITTI test set.} 
We present the results of pedestrians and cyclists on the test set of KITTI in Table \ref{table_2}. MonoLSS outperforms all image-only methods by a large margin. When compared with methods using extra data, MonoLSS performs better than most of them while only slightly weaker than CMKD \cite{YuH-CMKD-ECCV2022}. 

\noindent
\textbf{Results on Waymo val set.}
We evaluated our MonoLSS method on the Waymo dataset, which is more diverse than KITTI. The experimental results presented in Table \ref{table_waymo} demonstrate that MonoLSS achieves superior performance compared to the state-of-the-art DEVIANT method \cite{kumar2022deviant} across multiple evaluation metrics and thresholds, particularly for nearby objects. Notably, MonoLSS also outperforms PatchNet \cite{Ma_2020_ECCV} and PCT \cite{wang2021progressive} without utilizing depth information. Although MonoLSS's performance is slightly lower than that of CaDDN \cite{Reading_2021_CVPR}, it is worth noting that CaDDN relies on LiDAR data during training, whereas MonoLSS is an image-only approach.

\begin{figure*}
\centering
\includegraphics[width=1.0\linewidth]{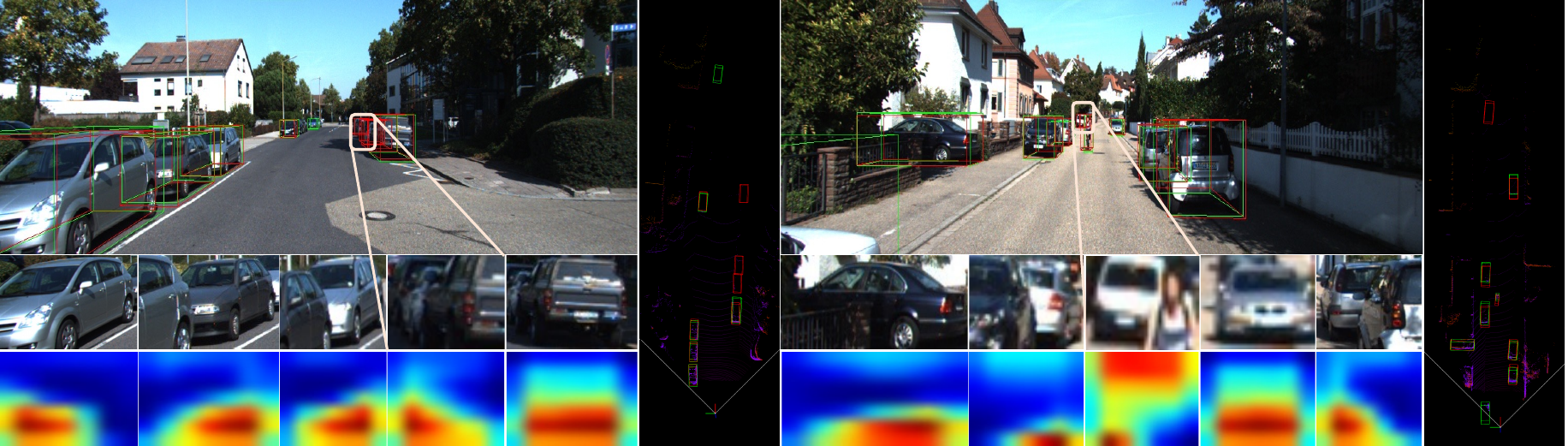}
\caption{\textbf{Qualitative visualization of some samples on KITTI \textit{val} set.} The 3D red boxes are produced by MonoLSS and the green boxes are the ground truth. Some unlabeled objects detected by MonoLSS are highlighted on images. The last line represents the LSS sampling map of the corresponding object. Best viewed in color with zoom in.}
\vspace{-5pt}
\label{show}
\end{figure*}

\noindent
\textbf{Cross-Dataset Evaluation.} 
Tabel \ref{table_cross} shows the result of our KITTI val model on the KITTI val and nuScenes \cite{caesar2020nuscenes} frontal val images, using mean absolute error (MAE) of the depth \cite{shi2021geometry}. MonoLSS is better than GUPNet \cite{Lu_2021_ICCV} and achieves similar competitive performance to DEVIANT \cite{kumar2022deviant}. This is because DEVIANT is equivariant to the depth translations and is more robust to data distribution changes.

\begin{table}
\begin{center}
\footnotesize
\setstretch{0.9}
\tabcolsep=0.65em
\begin{tabular}{ccc|c|c|ccc}
\hline
\multicolumn{1}{c}{\multirow{2}{*}{$D$}} & \multirow{2}{*}{$S$} & \multirow{2}{*}{$\theta$} & \multirow{2}{*}{$W$} & \multirow{2}{*}{$M$} & \multicolumn{3}{c}{$AP_{3D}/AP_{BEV}(IOU=0.7|{R_{40}})$}                                                   \\ \cline{6-8} 
\multicolumn{1}{c}{}                  &                       &                &                       &                       & \multicolumn{1}{c}{Easy} & \multicolumn{1}{c}{Mod.} & \multicolumn{1}{c}{Hard} \\ \hline\hline
      &    &    &  &   &21.72/30.74   &15.63/22.74  &12.80/19.14                     \\
\checkmark  & &   &      &     & 17.03/24.40    & 12.77/18.73        &  11.08/15.84                        \\
\checkmark  &   & &\checkmark   &  & 24.78/33.32 &17.65/23.92  &14.53/20.21 \\
\checkmark  &\checkmark  & \checkmark  &\checkmark  & &  24.63/33.63   & 17.55/25.03   & 14.62/21.48                            \\
 & & & & \checkmark & 24.65/34.15 & 17.33/24.56   & 14.34/20.97                            \\
\checkmark  &\checkmark &\checkmark   &\checkmark      &\checkmark     & 25.69/33.54    &  17.84/24.62   & 15.62/21.25 \\
\checkmark & &  &\checkmark  &\checkmark & \textbf{25.91}/\textbf{34.70} & \textbf{18.29}/\textbf{25.36} & \textbf{15.94}/\textbf{21.84} \\ \hline
\end{tabular}
\end{center}
\vspace{-15pt}
\caption{\textbf{Ablation Study on different components of our overall framework on KITTI \textit{val} set for Car category}. $D$, $S$, and $\theta$ denote that the LSS module acts on depth, dimension, and orientation angle, respectively. $W$ denotes warm-up strategy and $M$ denotes the MixUp3D.}
\vspace{-5pt}
\label{ablation_1}
\end{table}

\begin{table}
\begin{center}
\footnotesize
\setstretch{0.9}
\tabcolsep=1.5em
\begin{tabular}{c|ccc}
\hline
\multicolumn{1}{c|}{\multirow{2}{*}{Method}} & \multicolumn{3}{c}{$AP_{3D}/AP_{BEV}(IOU=0.7|{R_{40}})$} \\ \cline{2-4} 
\multicolumn{1}{c|}{}                          & \multicolumn{1}{c}{Easy}     & \multicolumn{1}{c}{Mod.}    & \multicolumn{1}{c}{Hard}    \\ \hline\hline
1$\times$1        &18.14/25.54  &12.68/18.68  &10.36/15.66     \\
3$\times$3        & 20.06/28.30 &15.28/21.95  &12.92/19.00     \\
5$\times$5        & 20.75/30.23 &15.14/22.39  &13.02/18.93     \\
7$\times$7    &21.72/30.74   &15.63/22.74      &12.80/19.14        \\
Depth       & 23.85/31.36     & 15.92/21.17    &11.64/16.76    \\
Seg       & 24.11/31.82     & 15.82/21.51    &13.09/16.90    \\
LSS       &  \textbf{24.78}/\textbf{33.32} & \textbf{17.65}/\textbf{23.92}  & \textbf{14.53}/\textbf{20.21} \\
\hline
\end{tabular}
\end{center}
\vspace{-15pt}
\caption{\textbf{Comparison of the LSS module with other sample selection methods on KITTI \textit{val} set for Car category.} The $\mathbf{1\times1}$, $\mathbf{3\times3}$, $\mathbf{5\times5}$ respectively represent that $1$, $9$, and $25$ sample points in the center are regarded as positive samples, and $\mathbf{7\times7}$ indicates that the method takes all sample points as positive samples. \textbf{Depth} denotes using extra depth map to select positive samples. \textbf{Seg} means using segmentation pseudo-labels generated by SAM \cite{kirillov2023segany}. \textbf{LSS} means our proposed LSS method.}
\vspace{-5pt}
\label{ablation_2}
\end{table}

\subsection{Ablation Study}
In this subsection, we investigate the impact of each component in our method\footnote{See the Supplementary Material for more ablation study, such as different Sampling Strategies on the LSS and influence of the MixUp3D.}. All ablation results are reported on the Car class of KITTI validation set and trained 150 epochs. To ensure result reliability, we report the median performance across five different seeds for each ablation experiment.

\noindent
\textbf{Effectiveness of the LSS Module.} As reported in the first 4 rows of Table \ref{ablation_1}, 
the LSS module can significantly improve the AP. While the effect is slightly off when it acts on the orientation and dimension property. This is because the depth estimation error is the most critical limiting factor in monocular 3D detection, which has been identified in GeoAug \cite{Lian_2022_CVPR} and MonoDLE \cite{Ma_2021_CVPR}. Thus, for the convenience of comparison, the LSS module only acts on the depth property in the following ablation experiments. 

\noindent
\textbf{Necessity of the warm-up.} Results in the first 3 rows of Table \ref{ablation_1} show the importance of the warm-up strategy. Without warm-up (2nd row), LSS will start random sampling at the beginning of training, leading to the fact that the true negative samples may be forced to learn attributes while the true positive samples are discarded instead, which significantly reduces the performance (21.72 to 17.03 in Easy level).

\noindent
\textbf{Effect between LSS and MixUp3D.} As presented in Table \ref{ablation_1}, while the LSS module and MixUp3D each exhibit a positive impact when applied independently ($+3.06$ and $+2.93$ in Easy level), their combined usage results in a higher improvement ($+4.19$).

\noindent
\textbf{Comparison of Sample Selection Strategies.} We contrast the LSS module with other sample allocation strategies, and the results are shown in Table \ref{ablation_2}. The LSS module adaptively selects positive samples based on object features, leading to a significant improvement in AP over methods that treat fixed-position sample points as positive ones. Furthermore, our method also outperforms the approach of using an additional depth or segmentation map to select positive samples.

\subsection{Qualitative Results}
In this subsection, we show 3D detection results of the MonoLSS in images and BEV maps. As shown in Figure \ref{show}, the MonoLSS can accurately estimate the 3D position of objects, even those not labeled by the annotators. In order to explore what features the LSS module is concerned with, we visualize the sampling map. As shown in the last row of Figure \ref{show}, the LSS module can adaptively determine the positive samples that are more suitable for learning 3D properties. Generally, it prefers to choose the bottom part of an object as positive samples. When occlusion occurs, the LSS module focuses on the regions without occlusion.
\section{Conclusion}

In this paper, we point out that the 3D property learning of monocular 3D detection faces a sample selection problem. We adopt a LSS module to adaptively determine the positive samples for each object. 
Moreover, we propose an unambiguous data augmentation method MixUp3D to improve the diversity of object-level samples.
Extensive experiments on the KITTI, Waymo and nuScenes benchmarks verify the effectiveness and efficiency of our MonoLSS. 
{
    \small
    \bibliographystyle{ieeenat_fullname}
    \bibliography{main}
}

\end{document}